\documentclass[11pt, a4paper, logo, copyright, nonumbering]{openreasonerzero}

\usepackage[utf8]{inputenc} %
\usepackage[T1]{fontenc}    %
\usepackage{hyperref}       %
\usepackage{url}            %
\usepackage{booktabs}       %
\usepackage{amsfonts}       %
\usepackage{nicefrac}       %
\usepackage{microtype}      %
\usepackage{xcolor}         %
\usepackage{comment}
\usepackage{fancyvrb}
\usepackage{listings}
\usepackage{algorithm}
\usepackage{algorithmicx}
\usepackage{algpseudocode}
\usepackage{subcaption}
\usepackage{cancel}
\usepackage{multirow}
\usepackage{amsmath}
\usepackage{graphicx}       %
\usepackage{CJKutf8}

\title{PaCoRe: Learning to Scale Test-Time Compute with Parallel Coordinated Reasoning}
\author[*]{
\vspace{-2ex}
{
\small
Jingcheng Hu$^{1,2*}$,~Yinmin Zhang$^{\footnotesize{1}}$, 
Shijie Shang$^{1}$, Xiaobo Yang$^{1,3*}$, 
Yue~Peng$^{1}$, Zhewei~Huang$^{1}$, Hebin~Zhou$^{1}$, Xin~Wu$^{1}$, 
Jie~Cheng$^{1}$, 
Fanqi~Wan$^{1}$, 
Xiangwen~Kong$^{1}$, 
Chengyuan~Yao$^{1}$, 
Kaiwen~Yan$^{1}$,
Ailin~Huang$^{1}$,
Hongyu~Zhou$^{1}$, 
Qi~Han$^{1}$, Zheng~Ge$^{1}$, Daxin~Jiang$^{1}$, Xiangyu~Zhang$^{1}$, 
Heung-Yeung Shum$^{2}$ 
}
\\
\footnotesize
\vspace{-1ex}
$^1$StepFun, $^2$Tsinghua University, $^3$Peking University \\
\begin{center}
\footnotesize
GitHub: \url{https://github.com/stepfun-ai/PaCoRe} \\
\footnotesize
Data: \url{https://huggingface.co/stepfun-ai/PaCoRe-Train-8k} \\
\footnotesize
Model: \url{https://huggingface.co/stepfun-ai/PaCoRe-8B} \\
\end{center}
\vspace{-5ex}
}
\correspondingauthor{Work done during internship at StepFun.}

\begin{document}

\begin{abstract}
We introduce Parallel Coordinated Reasoning (PaCoRe), a training-and-inference framework designed to overcome a central limitation of contemporary language models: their inability to scale test-time compute (TTC) far beyond sequential reasoning under a fixed context window.
PaCoRe departs from the traditional sequential paradigm by driving TTC through massive parallel exploration coordinated via a message-passing architecture in multiple rounds. 
Each round launches many parallel reasoning trajectories, compacts their findings into context-bounded messages, and synthesizes these messages to guide the next round and ultimately produce the final answer.
Trained end-to-end with large-scale, outcome-based reinforcement learning, the model masters the synthesis abilities required by PaCoRe and scales to multi-million-token effective TTC without exceeding context limits. 
The approach yields strong improvements across diverse domains, and notably pushes reasoning beyond frontier systems in mathematics: an 8B model reaches 94.5\% on HMMT 2025, surpassing GPT-5’s 93.2\% by scaling effective TTC to roughly two million tokens.
We open-source model checkpoints, training data, and the full inference pipeline to accelerate follow-up work.
\end{abstract}

\maketitle

\begin{figure}[h]
  \centering
  \begin{subfigure}[b]{0.48\linewidth}
    \includegraphics[width=\linewidth]{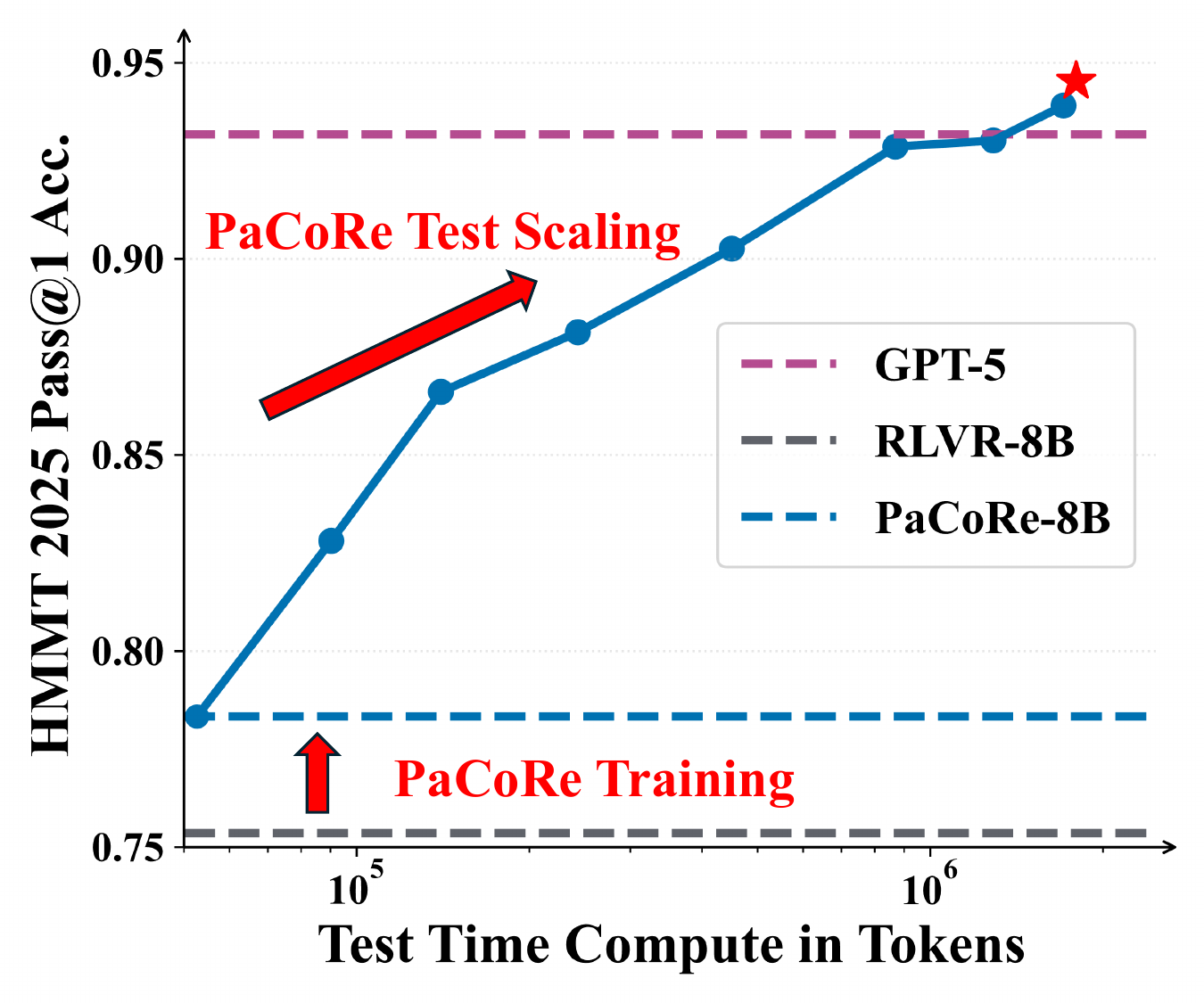}
  \end{subfigure}
  \hfill %
  \begin{subfigure}[b]{0.48\linewidth}
    \includegraphics[width=\linewidth]{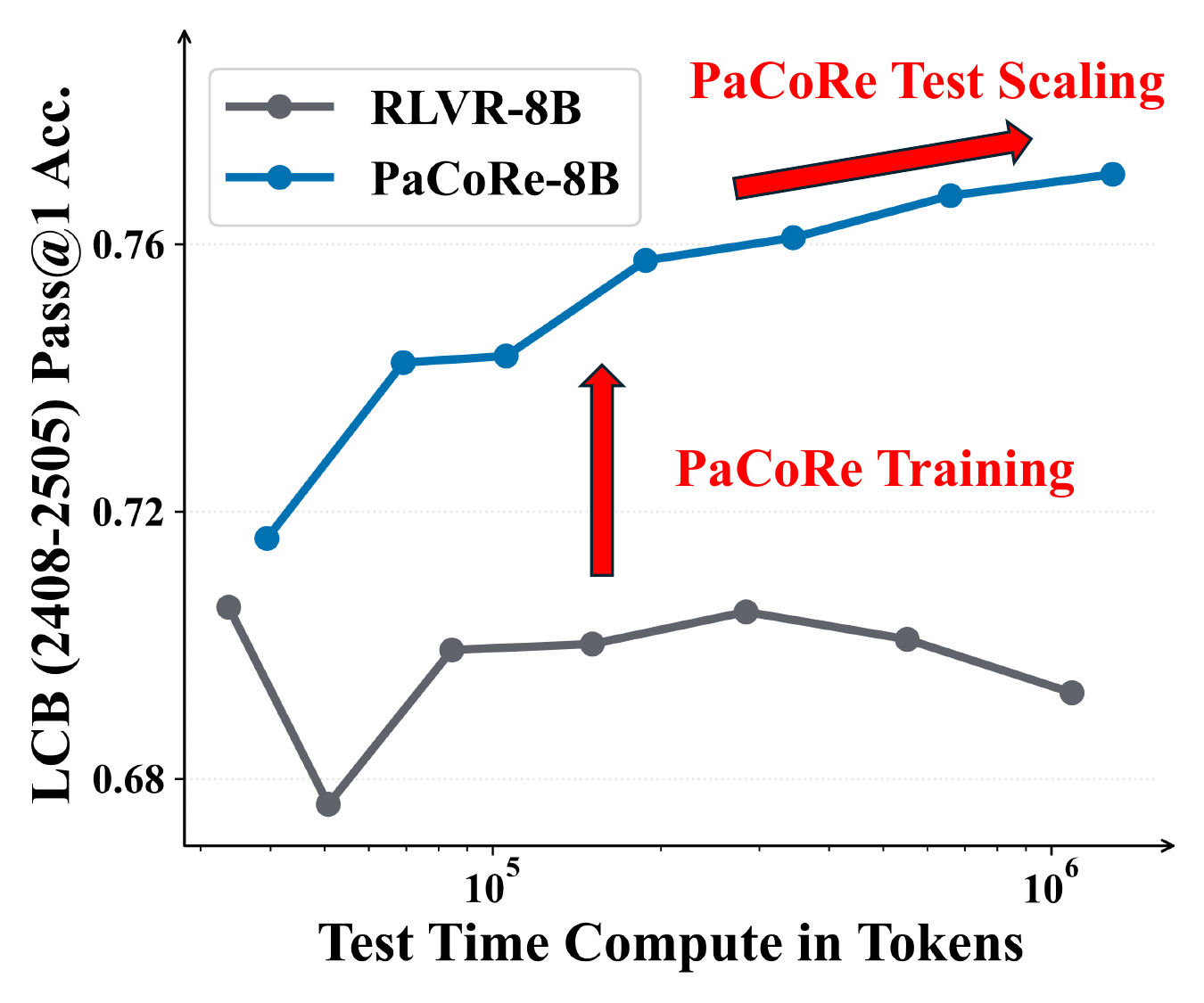}
  \end{subfigure}
  
  \caption{
  \textbf{Pa}rallel \textbf{Co}ordinated \textbf{Re}asoning (PaCoRe) performance.
\textbf{\emph{Left:}} On HMMT 2025, PaCoRe-8B demonstrates remarkable test-time scaling by increasing both parallel trajectories and coordinated rounds, yielding steady gains and ultimately surpassing GPT-5.
\textbf{\emph{Right:}}
On LiveCodeBench, the RLVR-8B model fails to leverage increased test-time compute, while PaCoRe-8B model effectively unlocks this synthesis capability, yielding substantial gains as the test-time compute increases.
}
  \label{fig:teaser}
\end{figure}

\newpage
\tableofcontents
\newpage

\section{Introduction}
\label{sec:intro}
Long-horizon reasoning, requiring sustained exploration, cross-checking, and iterative self-correction, underpins the most demanding intellectual pursuits.
Approaching such capabilities in AI demands substantial test-time compute (TTC)—the computation devoted to solving a single challenging instance.

A recurring lesson in the history of deep learning is that progress on hard problems often comes from scaling test-time compute through search~\cite{silver2016mastering,Li_2022,trinh2024solving} \textemdash expanding both in depth (sequentially) and width (in parallel)\textemdash rather than relying on a single forward pass. 
Yet for contemporary language models~\cite{brown2020languagemodelsfewshotlearners}, a fixed context window places a hard ceiling on how much such search-driven TTC can be expressed: standard sequential reasoning~\cite{wei2023chainofthoughtpromptingelicitsreasoning} packs every intermediate state into a single expanding chain, strictly coupling reasoning volume to context capacity; once the window fills, the reasoning must stop.

To address this, we introduce \textbf{Parallel Coordinated Reasoning (PaCoRe)}, 
a framework that decouples test-time compute scaling from context limits by shifting the primary driver of inference from solely sequential depth to elevated coordinated breadth. 
PaCoRe iteratively coordinates parallel reasoning trajectories and compresses their insights into compact messages. 
The messages from the previous round are then synthesized within the model’s context window to guide subsequent exploration, and ultimately, produce the final answer.

Crucially, PaCoRe’s success hinges on shifting the model from naive aggregation and solitary problem-solving to active coordination. Vanilla reasoning models tend to rely on \textbf{simple heuristics} such as majority voting~\cite{zhao2025majorityrightrltraining} when problems admit a simple, easily comparable answer. 
However, on problems with more complex solution structures (\textit{e.g.}, code generation and theorem proving), they often exhibit \textbf{Reasoning Solipsism}: despite receiving rich insights from parallel branches, they ignore this context and attempt to solve the problem from scratch, wasting the accumulated test-time compute (shown in Figure ~\ref{fig:teaser}, right).
To overcome this, we employ large-scale, outcome-based reinforcement learning, compelling the model to master \textbf{Reasoning Synthesis}: the capacity to scrutinize parallel branches, reconcile conflicting evidence, and synthesize a unified solution that exceeds any individual trajectory.

We evaluate PaCoRe across challenging benchmarks and observe substantial gains from massively scaling TTC far beyond the limits of standard decoding.
The results are especially pronounced in mathematics: on the HMMT 2025 competition benchmark, our PaCoRe-8B model achieves a score of 94.5\%, surpassing the leading proprietary reasoning model, GPT-5, at 93.3\%.
This performance is achieved by scaling effective TTC up to nearly two million tokens per problem, orchestrated entirely within the model's standard context window. 

We summarize our primary contributions as follows:
\begin{enumerate}
    \item We introduce \textbf{PaCoRe}, a general framework that decouples reasoning volume from model context capability by coordinating parallel reasoning, enabling multi-million token effective test-time compute. 
    \item We demonstrate that large-scale, outcome-based reinforcement learning is essential for inducing the synthesis capabilities required by this framework in diverse and challenging reasoning settings.
    \item We release comprehensive resources including model checkpoints, training data, and the inference pipeline, to enable rigorous evaluation and foster further research within the community.
\end{enumerate}

\begin{figure}[ht]
    \centering
    \includegraphics[width=\linewidth]{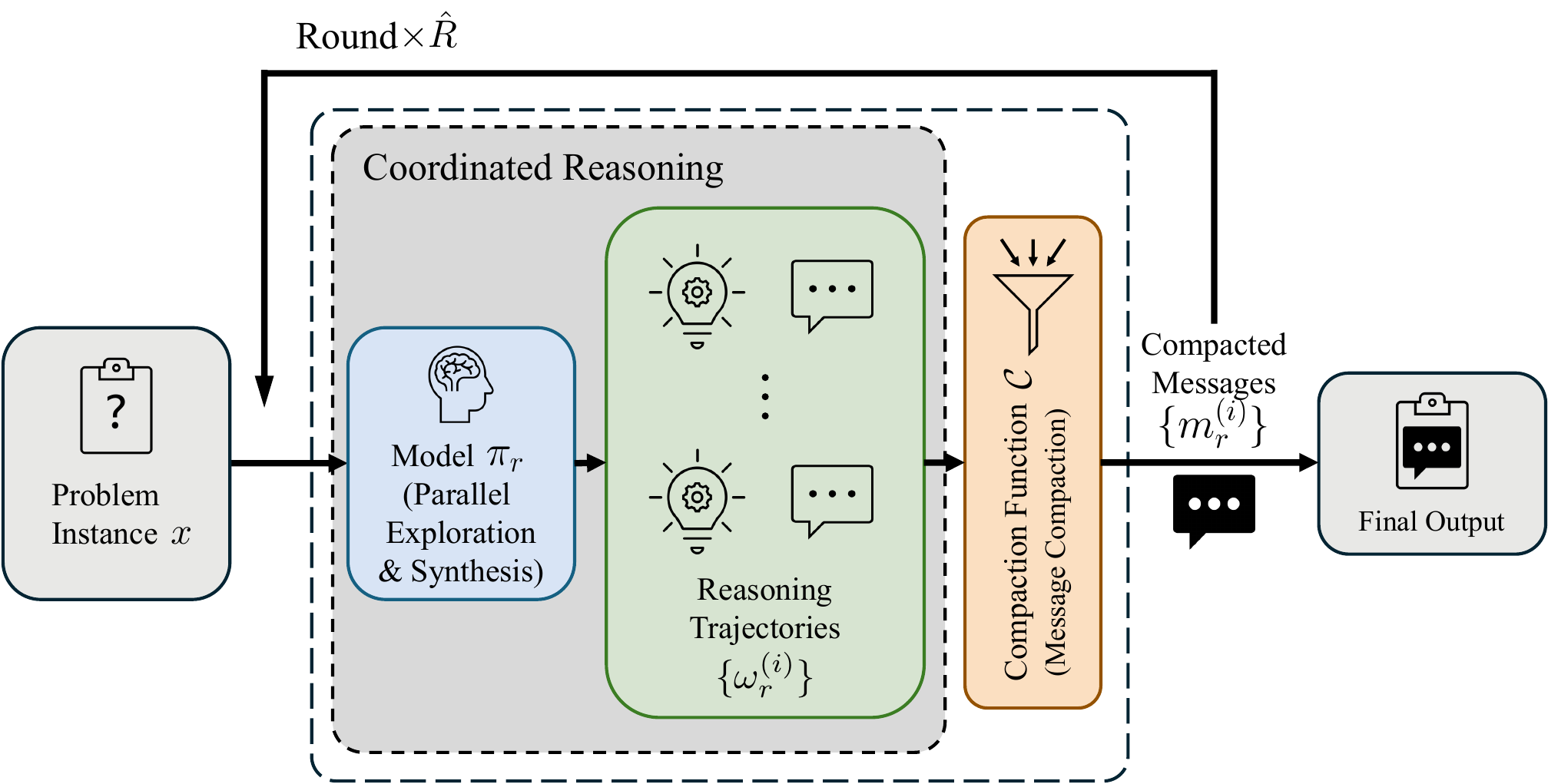}
    \caption{
    \textbf{Inference pipeline of PaCoRe.}
Each round launches broad parallel exploration, compacts the resulting trajectories into compacted messages, and feeds these messages together with the question forward to coordinate the next round. 
Repeating this process $\hat{R}$ times yields multi-million-token effective TTC while respecting fixed context limits, with the final compacted message serving as the system’s answer.
}
\end{figure}

\newpage

\section{Methods}
The framework comprises training and inference pipelines.
The inference pipeline coordinates parallel exploration to massively scale test-time compute, growing far beyond the limits of sequential reasoning while remaining independent of context constraints. 
The training procedure employs large-scale, outcome-based reinforcement learning to teach the model the synthesis skills required to consolidate diverse trajectories and generate high-quality final answers.

\subsection{Inference Pipeline}
Given a problem instance $x$, the PaCoRe inference pipeline executes $R$ rounds of \textbf{coordinated reasoning}. 
At each round $r \in \{1, \ldots, R\}$, the system inherits a set of \textbf{compact messages} 
$M_{r-1} = \{m_{r-1}^{(i)}\}_{i=1}^{K_{r-1}}$
from the previous round, 
and generates a new set of \textbf{reasoning trajectories} 
$\Omega_r = \{\omega_r^{(i)}\}_{i=1}^{K_r}$ that explore the solution space in parallel. 
The final answer is simply the degenerate case where $K_r=1$ at the last round.
For convenience, we refer to the preceding coordinated-reasoning process as consisting of $\hat R = R-1$ rounds, and refer $\vec{K} = [K_1,\ldots,K_{\hat{R}}]$ as inference trajectory configuration.

 For each problem $x$ , the model begins each round by taking the combined context $(x, M_{r-1})$ and generating a set of parallel trajectories $\Omega_r = \{\omega_r^{(i)}\}$. The message set $M_{r-1}$ provides a compacted summary of reasoning trajectories from the previous round, and each trajectory $\omega_r^{(i)}$ contains a full chain of reasoning followed by a final conclusion.
Each round then comprises two stages:
(i) \textbf{Synthesis and Parallel Exploration}, where the model takes the combined context $(x, M_{r-1})$ and produces the trajectories $\Omega_r$; and 
(ii) \textbf{Message Compaction}, where the trajectories $\Omega_r$ are compressed into the next-round message set $M_r$, allowing PaCoRe to massively scale effective test-time compute under a fixed context window.
We next detail these two stages of \textbf{coordinated reasoning}, outlining how PaCoRe constructs inputs, generates trajectories, and compacts them across rounds.

\begin{figure*}[t!]
  \centering
  \begin{subfigure}[b]{0.48\linewidth}
    \includegraphics[width=\linewidth]{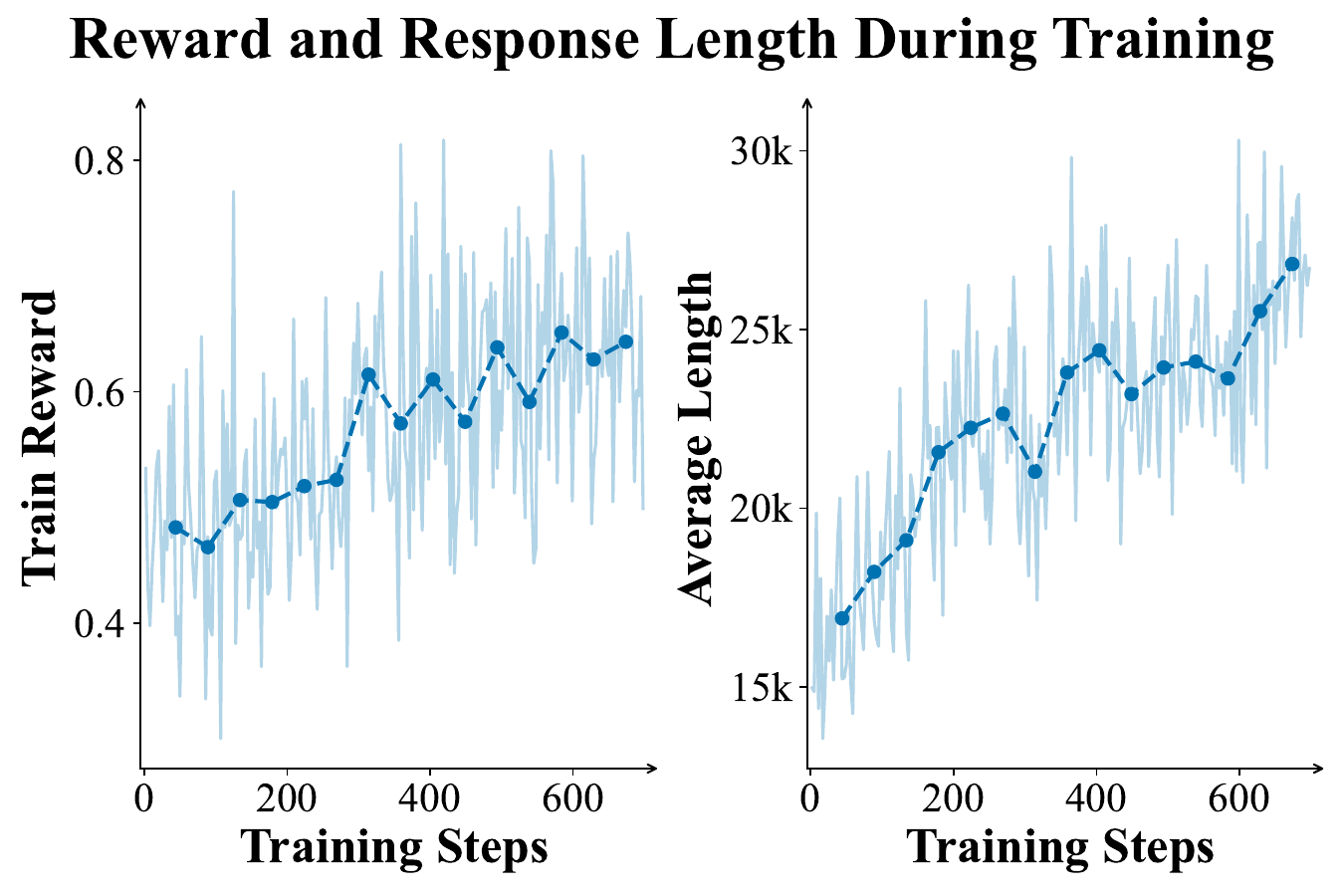}
    \label{fig:training_reward_response}
  \end{subfigure}
  \hfill %
  \begin{subfigure}[b]{0.48\linewidth}
    \includegraphics[width=\linewidth]{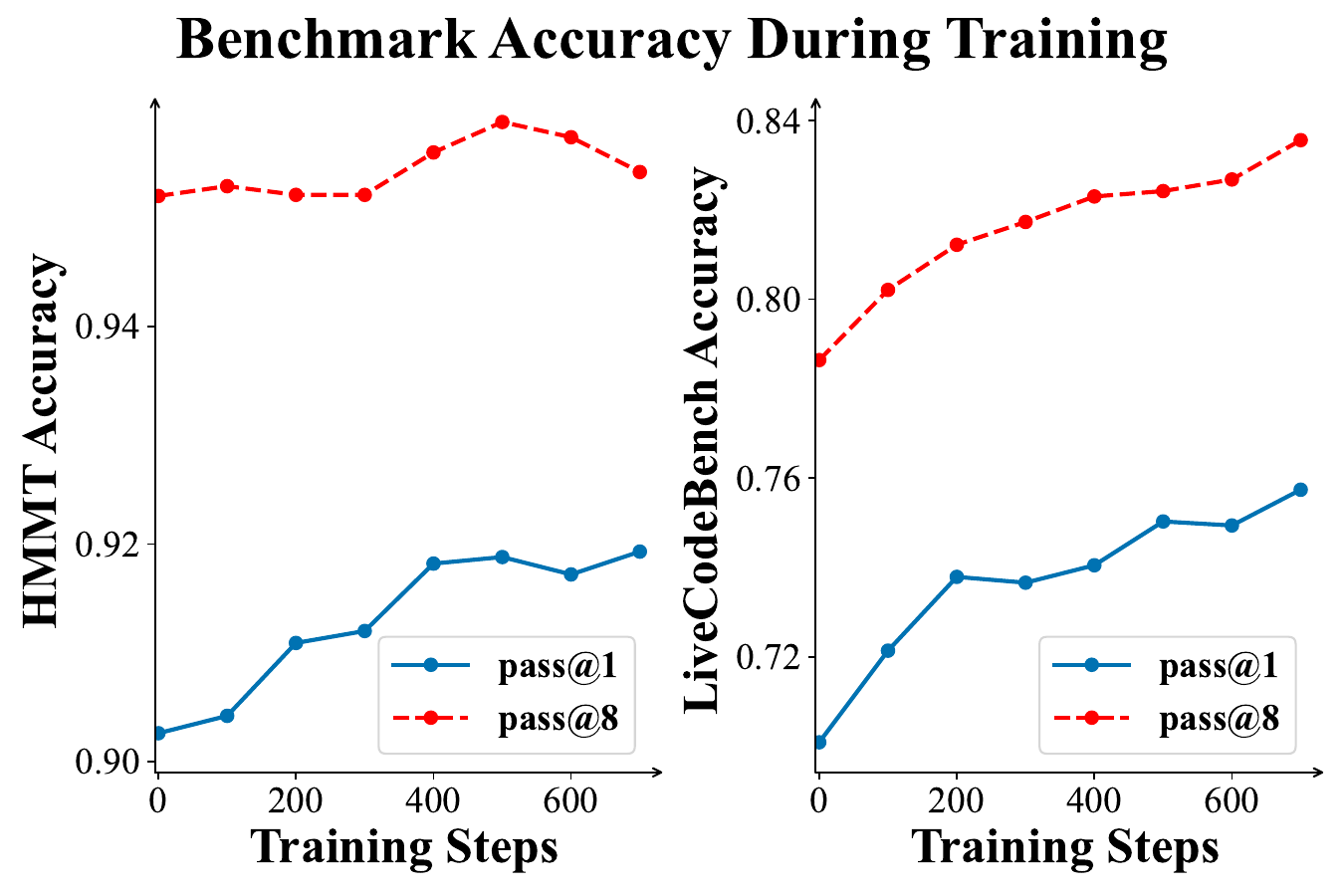}
    \label{fig:training_benchmark}
  \end{subfigure}
\vspace{-3ex}
    \caption{
    \textbf{PaCoRe Training dynamics.} 
    \textbf{\textit{Left panels:}} The Training Reward and Response Length steadily increase, demonstrating the training stability and effectiveness. 
    \textbf{\textit{Right panels:}} Evaluation on HMMT 2025 and LiveCodeBench (2408-2505). 
    Performance is reported using single round coordinated reasoning in PaCoRe inference setting with $\vec{K} = [16]$. 
    }
    \label{fig:train-time-scale}
  \label{fig:both}
\end{figure*}

\paragraph{Synthesis and Parallel Exploration.}
At the beginning of round $r$, the system uses the prompting function $P(x, M_{r-1})$ to serialize the problem and the compact messages, producing a structured input sequence for the model.

Once the input is constructed, the core reasoning model~$\pi_r$ is then invoked in parallel to produce $K_r$ independent trajectories:
\begin{gather}
    \omega_r^{(i)} \sim \pi_r(\,\cdot \mid P(x, M_{r-1})\,).
\end{gather}
In our implementation, we employ the same model weights across all rounds for operational convenience. 
This parallel generation constitutes the main drive of effective TTC:
even though the system can accumulate millions of tokens through parallel expansion, each round’s input consumes only $(x, M_{r-1})$, maintaining an almost constant context cost.

\paragraph{Message Compaction.}
To enable multi-round coordination while respecting the fixed context window, the full trajectories in $\Omega_r$ must be compressed into a small set of messages. We apply a compaction function:
\begin{gather}
    M_r = \mathcal{C}(\Omega_r),
\end{gather}
where $\mathcal{C}$ transforms the set of trajectories into a new set of compact messages.

In our implementation, we leverage the structured nature of the generated trajectories. 
Reasoning models typically produce detailed intermediate derivations (\textit{e.g.}, chain-of-thought or "reasoning content") followed by a final conclusion.
We therefore implement $\mathcal{C}(\cdot)$ as a trajectory-wise extraction function $C(\cdot)$: it parses each $\omega_r^{(i)} \in \Omega_r$, retains only the final conclusion segment (forming $ m_r^{(i)} = C(\omega_r^{(i)}).$), and discards the intermediate steps.
This ensures that the input length required for synthesis remains bounded (\textit{i.e.}, $|x| + \sum_{i} |m_{r}^{(i)}|$ fits within the context window), even as the effective TTC\textemdash aggregate token count across all trajectories, $\sum_r \sum_i|\omega_r^{(i)}|$\textemdash scales vastly.

\paragraph{Iterative Coordination.}
The process repeats for all $R$ rounds, with compact messages progressively refining the model’s understanding of the problem.
To ensure convergence, the final round uses a single trajectory ($K_R = 1$), 
producing a final compact message
\begin{gather}
    y = m_R^{(1)},
\end{gather}
which constitutes the output of the PaCoRe inference pipeline.

\begin{table}[t]
    \centering
    \resizebox{0.99\linewidth}{!}{
    \begin{tabular}{lccccccc}
        \toprule
        \multirow{2}{*}{\textbf{Benchmark}} & \textbf{AIME} & \textbf{HMMT} & \textbf{IMO} & \multirow{2}{*}{\textbf{Apex}} & \multirow{2}{*}{\textbf{LiveCodeBench}} & \multirow{2}{*}{$\textbf{HLE}_{\text{text}}$} & \textbf{Multi} \\
        & \textbf{2025} & \textbf{2025} & \textbf{AnswerBench} & & & & \textbf{Challenge} \\
        \midrule
        GPT-5 & 93.5 (13k) & 93.2 (16k) & 72.9 (26k) & 1.0 (33k) & \textbf{83.5} (13k) & \textbf{26.0} (14k) & \textbf{71.1} (5.0k) \\
        Qwen3-235B-Thinking & 91.6 (26k) & 82.3 (32k) & 71.7 (34k) & \textbf{3.3} (46k) & 74.5 (21k) & 18.2 (23k) & 60.3 (1.6k) \\
        GLM-4.6 & 92.3 (20k) & 88.7 (25k) & 73.5 (37k) & 0.7 (53k) & 79.5 (19k) & 17.2 (21k) & 54.9 (2.2k) \\
        DeepSeek-v3.1$^{*}$ & 90.2 (16k) & 86.1 (20k) & 63.0 (27k) & 1.4 (36k) & 74.9 (11k) & 19.3 (18k) & 54.4 (1.1k) \\
        Kimi-K2-Thinking & \textbf{95.3} (25k) & 86.5 (33k) & 76.5 (44k) & 0.8 (60k) & 79.2 (25k) & 23.9 (29k) & 66.4 (1.6k) \\
        \midrule
        RLVR-8B & 84.1 (50k) & 75.4 (48k) & 64.6 (56k) & 0.0 (65k) & 70.6 (34k) & 9.3 (35k) & 33.3 (1.7k) \\
        \textbf{PaCoRe-8B} (low) & 89.7 (255k) & 88.1 (243k) & 76.1 (306k) & 0.7 (362k) & 75.8 (188k) & 13.0 (196k) & 41.8 (13k) \\
        \textbf{PaCoRe-8B} (medium) & 92.5 (908k) & 92.9 (869k) & 77.3 (1080k) & 1.4 (1280k) & 76.7 (659k) & 14.6 (694k) & 45.7 (45k) \\
        \textbf{PaCoRe-8B} (high) & 93.7 (1873k) & \textbf{94.5} (1796k) & \textbf{78.4} (2258k) & 2.3 (2679k) & 78.2 (1391k) & 16.0 (1451k) & 48.0 (95.3k) \\
        \bottomrule
    \end{tabular}
    }
    \caption{
    \textbf{Benchmark performance and TTC spent per problem instance.}
    For each benchmark, we report accuracy together with total TTC (in thousands).
    $^{*}$DeepSeek-V3.1 refers to the Terminus version.
    For \textit{Low}, \textit{Medium}, and \textit{High}, we apply the inference configuration as $\vec{K}=[4]$, $[16]$, and $[32, 4]$ separately.
    }
    \label{tab:main_table}
    \vspace{-3ex}
\end{table}

\subsection{Training Procedure}
The efficacy of the PaCoRe framework fundamentally depends on the core model's capacity for \textbf{\textit{synthesis}}: the ability to critically evaluate diverse perspectives within the input messages $M$, reconcile conflicting information, and generate novel strategies that surpass the quality of any individual input.

To train this capability, we instantiate the synthesis stage of a single PaCoRe round as an episodic RL environment, with the core model serving as the policy to be optimized. 
In each training episode, we sample a problem $x$ and a corresponding set of input messages $M$ from a training distribution $D$, where M can be either generated online or drawn from a cached pool. 
Given the sampled pair $(x, M)$, the policy $\pi_\theta$ generates a reasoning trajectory $\omega$ from the formatted input $P(x, M)$, and receives a sparse terminal reward $R(\omega) \in [0,1]$ at the end.

We then apply large-scale, outcome-based reinforcement learning to elicit these advanced synthesis behaviors.
Training is performed on challenging domains, where each trajectory is evaluated by the correctness of its extracted message. 
In order to ensure that naive strategies like majority voting or random selection are insufficient and compel the policy to develop synthesis to succeed, we discard training instances where the average accuracy of the message set $M$ exceeds a predefined threshold.

Compared to standard RL on static tasks with verifiable rewards \cite{lambert2025tulu3pushingfrontiers}, PaCoRe represents a significant departure. 
Because the input context $M$ is composed of model-generated messages, the policy must function within an implicitly multi-agent environment rather than a fixed single-agent setting. 
Although the outcome-based reward aligns with conventional RL formulations for chain-of-thought optimization, PaCoRe demands higher-level synthesis and coordination across agents' outputs, introducing opportunities to examine emergent collective behaviors.

\begin{figure*}[t]
    \centering
    \includegraphics[width=0.95\linewidth]{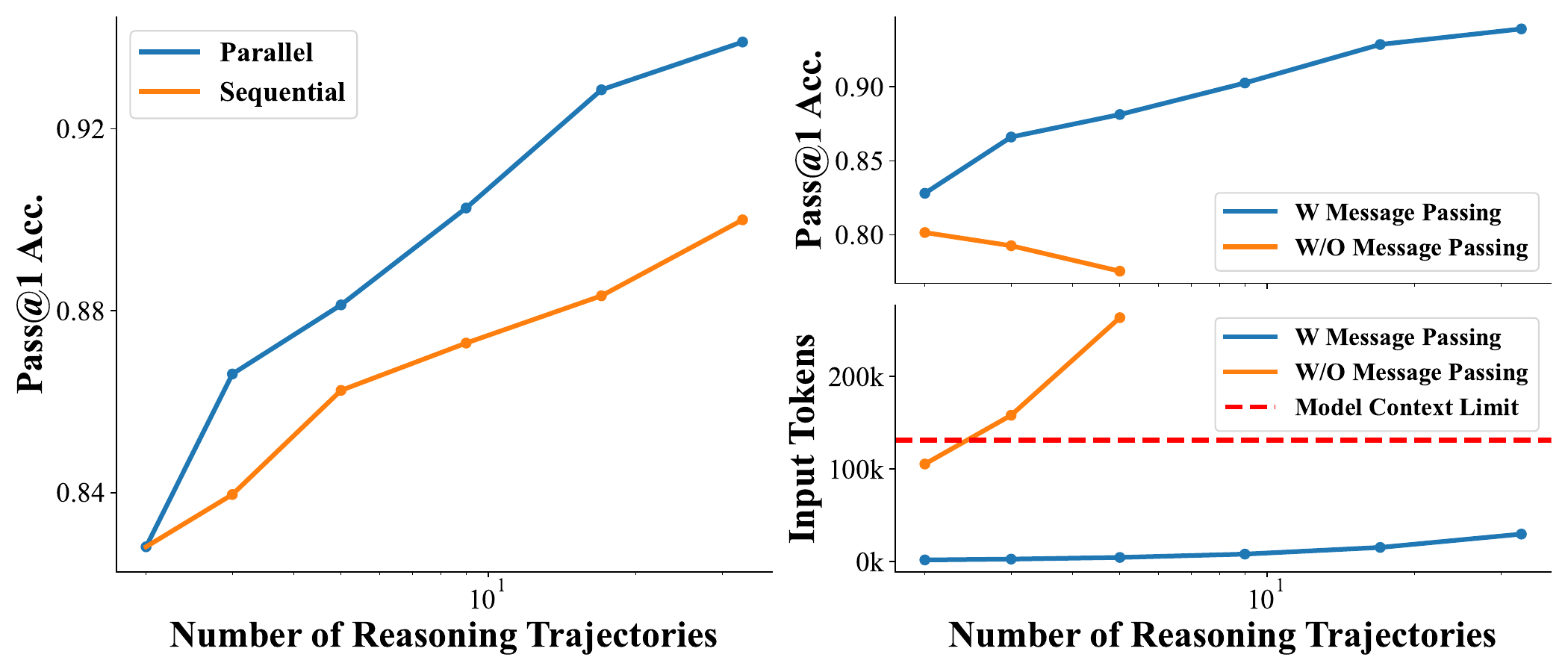}
       \caption{
    \textbf{Ablation of parallel reasoning and message passing.} 
    \textbf{\textit{Left:}} \textbf{Parallel} scaling ($\vec{K} = [N, ]$) utilizes test-time compute more effectively than \textbf{Sequential} scaling ($\vec{K} = [1, \ldots, 1]$).
    \textbf{\textit{Right:}} \textbf{Message Passing} is essential for test-time scaling. Without compaction ("W/O Message Passing"), performance degrades as test time scales and fundamentally limited by model context length, whereas standard PaCoRe ("W Message Passing") scales unboundedly and robustly. Pass@1 accuracy is evaluated on HMMT 2025.
    } 
    \label{fig:key-design-ablation}
\end{figure*}

\section{Experiments}
In this section, we present comprehensive experimental results and analysis of our PaCoRe framework. 
We begin by detailing the training protocol used to develop our PaCoRe-8B model, followed by in-depth analysis of training results. 
We then present the main evaluation results, comparing PaCoRe against frontier reasoning models on challenging benchmarks. 
Finally, we conduct analysis and ablation studies to validate key design choices and investigate the properties of the learned synthesis capabilities.

\begin{figure*}[htbp]
    \centering
    \includegraphics[width=0.95\linewidth]{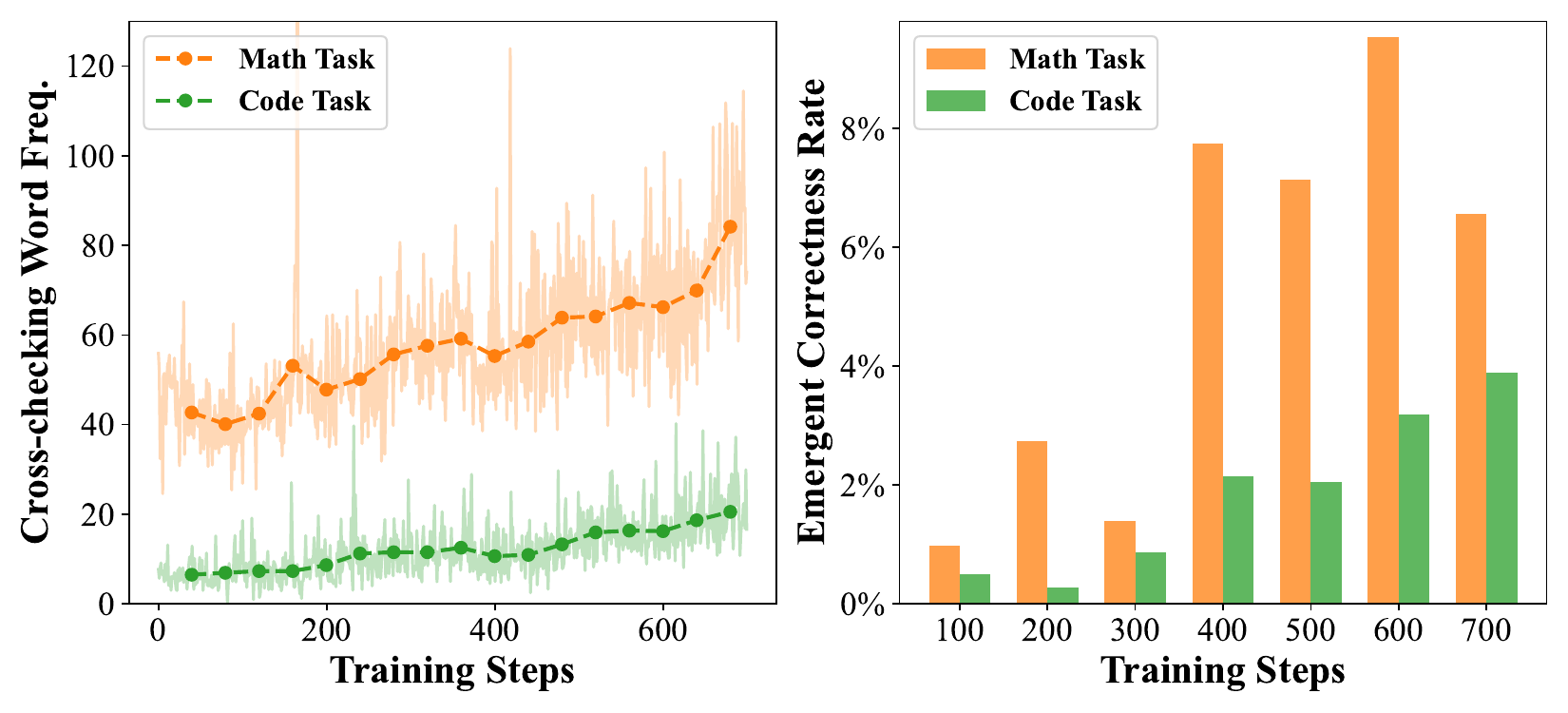}
    \caption{
    \textbf{Evolution of synthesis-related linguistic features and emergent correctness across training steps.}
    \textbf{\textit{Left:}} Frequency of cross-checking words (including 'reference', '\begin{CJK*}{UTF8}{gbsn}参考\end{CJK*}', 'Ref <number>', 'ref <number>') in generated solutions. 
    Training elicits and magnifies this capability across domains; notably, the near-zero initial frequency in Code corroborates the poor test-time scaling of untrained models (Figure~\ref{fig:teaser}).
    \textbf{\textit{Right:}} The \textbf{Emergent Correctness Rate} tracks the probability of generating a correct solution given input messages that are all incorrect, averaged over 100-step intervals. 
    The upward trend in both domains demonstrates that through large scale RL training, the model transcends naive strategies like majority voting or random selection to achieve genuine synthesis, recovering valid solutions even from entirely erroneous contexts.
    }
    \label{fig:linguistic_features}
\end{figure*}

\subsection{Training} 
\label{sec:training_details}

\paragraph{Training Recipe.}
We initialize PaCoRe-8B from an internal, reasoning-oriented post-trained version of Qwen3-8B-Base \cite{yang2025qwen3technicalreport} (denoted as RLVR-8B). 
We use RLVR-8B to generate a message cache pool of size 24 for each problem in the training dataset. 
During training, we randomly sample the message set $M$ from this pool with $|M| \sim U(16, 24)$. 
We employ strict on-policy PPO~\cite{schulman2017proximalpolicyoptimizationalgorithms} with GAE ($\lambda = 1, \gamma = 1$) following the ORZ setup~\cite{hu2025openreasonerzero}, utilizing a batch size of 16 instances (64 responses each), a maximum sequence length of 131,072, and generation temperature/top-p of 1.0.
We leverage competition-level mathematics and coding tasks as the primary substrate to cultivate model capabilities.
To incentivize the model to learn synthesis, the training data distribution is refined over two stages. 
In \textbf{Stage 1} (250 iterations), we target scenarios where naive aggregation fails by filtering for low message set accuracy ($0 < \text{mean}(\text{message\_acc}) < 9/24$ for math; $< 15/24$ for code), alongside quality filtering.
In \textbf{Stage 2} (450 additional iterations), we further refine the distribution. We use an intermediate Stage 1 checkpoint to evaluate synthesis accuracy on the Stage 1 data, retaining only instances where $0 < \text{synthesis\_acc} < 1$. 
The final PaCoRe-8B model is obtained after 700 total iterations.

\paragraph{Training Results.}
We present the training dynamics by tracking training rewards, response lengths, and performance on key benchmarks. 
As illustrated in Figure~\ref{fig:train-time-scale}, both the training reward and response length steadily increase throughout the process. 
Correspondingly, the performance on both HMMT 2025 and LiveCodeBench (2408-2505) improves significantly.
In these evaluations, the PaCoRe inference setting is $\vec{K} = [16,]$. In order to improve experimental efficiency, we utilize cached responses from the RLVR-8B model for each problem instance to directly seed the current checkpoint for the next round generation. 
All these results indicate stable and effective learning.

\subsection{Evaluation}
\label{sec:main_results}

\paragraph{Evaluation Setup.}
We evaluate PaCoRe on representative benchmarks across mathematics, coding, science, and open-ended generation: AIME 2025, HMMT 2025 (Feb.), Apex, IMOAnswerBench, LiveCodeBench~(2408-2505), Humanity's Last Exam (text subset), and MultiChallenge~\cite{balunovic2025matharena, luong2025robustmathematicalreasoning, jain2024livecodebench, phan2025humanitysexam, sirdeshmukh2025multichallengerealisticmultiturnconversation}.
We compare against frontier reasoning models including GPT-5 ("high reasoning effort) \cite{gpt5}, Kimi-K2-Thinking \cite{k2_thinking}, DeepSeek-V3.1-Terminus \cite{ds_v3p1_terminus}, GLM-4.6 \cite{glm_4p6}, and Qwen3-235B-A22B-Thinking-2507 \cite{qwen3_235b_a22b_thinking_2507}. 
We report pass@1 accuracy for all approaches based on average performance of multiple independent generations per problem: 64 for AIME 2025, HMMT 2025, and APEX; 8 for IMOAnswerBench, LiveCodeBench, and MultiChallenge; and 1 for HLE.
For both PaCoRe and RLVR evaluations, We maintain a maximum generation sequence length of 131,072 tokens with a temperature and top-p of 1.0, and we apply YaRN~\cite{peng2023yarnefficientcontextwindow} (scale 2.0) without modifying attention temperature. 
We evaluate our PaCoRe inference pipeline under three coordinated reasoning levels: \textit{Low} ($\vec{K}=[4]$), \textit{Medium} ($\vec{K}=[16]$), and \textit{High} ($\vec{K}=[32, 4]$).
To optimize experimental efficiency without compromising validity, we utilize a caching strategy for the first round for PaCoRe evaluation: we pre-generate a pool of 512 trajectories for each problem, from which the model randomly samples $K_1$ items to seed the next round. We empirically validated that this method yields results equivalent to generating all trajectories from scratch.

\begin{table*}[t]
    \centering
    \resizebox{0.9\linewidth}{!}{
    \begin{tabular}{llcccc}
        \toprule
        \multirow{2}{*}{\textbf{Method}} & \multirow{2}{*}{\textbf{Setting}} & \multicolumn{4}{c}{\textbf{Benchmark}} \\
        \cmidrule(lr){3-6}
         & & \shortstack{\textbf{AIME} \\ \textbf{2025}} & \shortstack{\textbf{HMMT} \\ \textbf{2025}} & \shortstack{\textbf{IMO} \\ \textbf{AnswerBench}} & \raisebox{1ex}{\textbf{Apex}} \\
        \midrule
        \multirow{4}{*}{\shortstack{\textbf{Self} \\ \textbf{Consistency}}} 
         & @4 & 87.0 (194k) & 82.3 (193k) & 69.7 (226k) & 0.3 (249k) \\
         & @16 & 88.5 (776k) & 85.9 (773k) & 72.1 (904k) & 0.0 (995k) \\
         & @64 & 89.5 (3106k) & 85.1 (3094k) & 72.8 (3615k) & 0.0 (3981k) \\
         & @256 & 90.0 (12422k) & 84.7 (12376k) & 73.0 (14458k) & 0.0 (15924k) \\
        \midrule
        \multirow{3}{*}{\textbf{PaCoRe}} 
         & low & 89.7 (255k) & 88.1 (243k) & 76.1 (306k) & 0.7 (362k) \\
         & medium & 92.5 (908k) & 92.9 (869k) & 77.3 (1080k) & 1.4 (1280k) \\
         & high & 93.7 (1873k) & 94.5 (1796k) & 78.4 (2258k) & 2.3 (2679k) \\
        \bottomrule
    \end{tabular}
    }
    \caption{
        \textbf{Test-time scaling comparison with Self-Consistency sampling (Majority voting).} 
    We compare PaCoRe against standard Self-Consistency (SC) on representative short-answer benchmarks. 
    Token counts (TTC) are reported in parentheses in thousands (k).
    Self-Consistency sampling are conducted using RLVR-8B and repeated sampling from a 512 size of pool of responses for each problem for 64 times to estimate the accuracy. 
    }
    \label{tab:cons_pacore_comparison}
\end{table*}

\paragraph{Evaluation Results.}
The main results are summarized in Table~\ref{tab:main_table}.  
Across all test-time effort settings, PaCoRe-8B consistently outperforms the starting checkpoint, RLVR-8B, on every benchmark.
Even in the \textit{Low} setting, PaCoRe-8B demonstrates robust capabilities, achieving 88.1\% on HMMT2025 and 75.8\% on LiveCodeBench, significantly outperforming the RLVR-8B baseline and surpassing considerably larger models like Qwen3-235B-A22B-Thinking-2507 and DeepSeek-V3.1-Terminus.
With further test-time scaling in the \textit{High} setting, PaCoRe-8B reaches a remarkable 94.5\% on HMMT 2025 and 78.4\% on IMOAnswerBench, surpassing the leading proprietary model, GPT-5, by scaling effective TTC to approximately 2 million tokens. 
A standout result is observed on the extremely challenging Apex benchmark: while the RLVR-8B fails to answer any questions correctly (0.0\%), PaCoRe-8B achieves 2.3\% in the \textit{High} setting. 
On LiveCodeBench, PaCoRe-8B achieves a strong 78.2\%, remaining competitive with frontier models like GLM-4.6 and Kimi-K2-Thinking.
These findings demonstrate that parallel coordinated reasoning allows an 8B model to bridge the gap with, and often surpass, state-of-the-art systems on complex reasoning tasks.

\subsection{Analysis and Ablations}
\label{sec:analysis}

\paragraph{Ablation of Core Design Principles.}
We study the importance of the two central components of PaCoRe: parallel exploration and message passing.
First, we compare parallel scaling (PaCoRe with $\vec{K} = [N, ]$) against sequential scaling (equivalent to $\vec{K} = [1, \ldots, 1]$) under the same total number of generated trajectories. 
As shown in Figure~\ref{fig:key-design-ablation} (Left), parallel coordinated reasoning utilizes TTC much more effectively than purely sequential approach.
Second, we examine the role of message passing. 
We compare PaCoRe (W Message Passing) with a variant where compaction is disabled (W/O Message Passing), feeding the full trajectories into the next round context. 
Figure~\ref{fig:key-design-ablation} (Right) shows that without compaction, performance degrades as TTC scales and is fundamentally limited by the context length. 
In contrast, PaCoRe, with message passing, scales robustly and without bound.

\paragraph{Evolution of Synthesis Capabilities.}
We probe the mechanism underlying PaCoRe by tracking the evolution of synthesis behaviors during training.
First, the frequency of "cross-checking" linguistic markers rises steadily across both domains (Figure~\ref{fig:linguistic_features}, Left).
Notably, the near-zero initial frequency in the Code domain aligns with the poor test-time scaling of untrained models (Figure~\ref{fig:teaser}), confirming that training fundamentally transforms the model into a coordinated reasoner.
To distinguish genuine synthesis from simple aggregation or refinement, we track the \textit{Emergent Correctness Rate} (Figure~\ref{fig:linguistic_features}, Right), defined as the probability of generating a correct solution given input messages that are all incorrect.
The upward trend in both domains demonstrates that the model transcends naive strategies like majority voting or random selection (which fail given all wrong inputs), learning instead to reconstruct valid solutions from erroneous partial evidence.
This capability is empirically confirmed in Table~\ref{tab:cons_pacore_comparison}, where PaCoRe demonstrates robust test-time scaling while the voting-based Self-Consistency baseline saturates rapidly.

\begin{table}[t]
    \centering
    \begin{tabular}{lc}
        \toprule
        \textbf{Model} & \textbf{SWE-Verified} \\
        \midrule
        RLVR-8B & 29.8\% \\
        PaCoRe-8B (low) & \textbf{34.0\%} \\
        \bottomrule
    \end{tabular}
    \caption{\textbf{Performance on SWE-Verified.} PaCoRe-8B (low) markedly surpasses the RLVR-8B baseline, indicating generalization to software engineering scenarios without task-specific tuning.}
    \label{tab:swe_verified}
\end{table}

\paragraph{Generalization Across Domains.}
To assess cross-domain generalization beyond mathematical and logical reasoning benchmarks, we evaluate PaCoRe on software engineering and multi-turn conversation tasks using the SWE-Verified~\cite{swe_verified} and MultiChallenge~\cite{sirdeshmukh2025multichallengerealisticmultiturnconversation}, respectively.
We note that the starting checkpoint (RLVR-8B) underwent no specialized post-training for these tasks; thus, this study assesses the intrinsic transfer of reasoning capabilities rather than fully optimized performance.
For SWE-Verified, we compare a compute-efficient variant, PaCoRe-8B (low), against the RLVR-8B baseline, both under Agentless~\cite{xia2024agentlessdemystifyingllmbasedsoftware} framework. 
As shown in Table~\ref{tab:swe_verified}, PaCoRe-8B (low) achieves a resolve rate of 34.0\%, substantially outperforming the baseline (29.8\%).
Furthermore, PaCoRe again exhibits strong generalization on MultiChallenge, with PaCoRe-8B (high) improving the RLVR-8B baseline from 33.3\% to 48.0\% under higher TTC budgets as shown in Table~\ref{tab:main_table}

\paragraph{General Effectiveness of PaCoRe Data. }
Beyond the framework itself, 
we discovered that the training corpus curated for PaCoRe constitutes a general applicable learning resource of exceptional density. 
As shown in Table~\ref{tab:30a3}, we observe that employing our released dataset as the primary substrate for standard RLVR yields a robust performance boost. 
The RLVR stage uses high-quality, carefully curated training data and applies an off-policy PPO algorithm with GAE ($\gamma=1, \lambda=1$) for only 50 iterations and 4 mini-batches per iteration on ours-SFT-Qwen3-30B-A3B. Despite this minimal compute budget, RLVR delivers substantial gains over the SFT model on both AIME 2025 and LiveCodeBench.
This indicates that our problem set—carefully filtered to demand genuine synthesis—serves as a highly effective catalyst for training strong reasoning models in general.

\paragraph{Ablation on Message Sampling Strategy in Training.}
We investigate the impact of message set size $|M|$ during training by comparing fixed versus uniformly sampled sizes.
As shown in Table~\ref{tab:sample_M_sizes}, training with diverse and relatively large message sets ($|M|\sim U(8,16)$) yields the best performance when evaluated under a fixed inference configuration ($\vec{K}=[16,]$).
This randomized sampling strategy improves robustness by familiarizing the model with varying parallel coordination settings, thereby facilitating superior test-time scaling.

\begin{table}[tbp]
    \centering
    \begin{tabular}{lcc}
        \toprule
        \textbf{Model} & \textbf{AIME 2025} & \textbf{LiveCodeBench} \\
        \midrule
        Ours-SFT-Qwen3-30B-A3B & 81.4\%  & 66.0\% \\
        + RLVR with PaCoRe Data & \textbf{83.2\%} & \textbf{74.0\%} \\
        \bottomrule
    \end{tabular}
    \caption{\textbf{General Effectiveness of PaCoRe Data.} RLVR trained on PaCoRe data, with only 200 update iterations, achieves significant gains on benchmarks}
    \label{tab:30a3}
    \vspace{-3ex}
\end{table}

\section{Related Work}
Classical approaches to sequentially scaling TTC, such as CoT \cite{wei2023chainofthoughtpromptingelicitsreasoning}, pack intermediate steps into a single expanding chain. 
Even when boosted by large-scale RL \cite{o1,deepseekai2025deepseekr1incentivizingreasoningcapability}, these methods inevitably saturate the context, strictly ceiled on achievable TTC. 

A shift towards parallel scaling has shown immense potential. Early methods \cite{wang2023selfconsistencyimproveschainthought,  trinh2024solving, openai2025competitiveprogramminglargereasoning} use simple rules to integrate language models with extensive parallel search, achieving strong performance in specific domains. 
However, the coordination in these systems often hinges on task-specific scaffolds or priors, hindering their general applicability.

This highlights the need for a general framework to coordinate massive parallel exploration within model context limits. 
Attempts at such coordination \cite{chen2023universalselfconsistencylargelanguage,  wang2024mixtureofagentsenhanceslargelanguage, wen2025parathinkernativeparallelthinking,
zheng2025parallelr1parallelthinkingreinforcement, venkatraman2025recursiveselfaggregationunlocksdeep,wu2025nativeparallelreasonerreasoning} synthesize parallel outputs but lack mechanisms for message passing or context management, thus facing the same context limitations. 
Alternative approaches focused on context management \cite{yan2025inftythinkbreakinglengthlimits,aghajohari2025markovianthinkerarchitectureagnosticlinear} use compression rules but remain rooted in the sequential paradigm, limiting TTC scaling efficiency. 
AggLM \cite{zhao2025majorityrightrltraining} focuses on learning aggregation strategies that surpass majority voting and reward-model ranking, while we introduce parallel coordinated reasoning with context compaction to scale test-time compute beyond context limits.
Thus, scaling trajectory configurations and rounds of coordinated reasoning yields substantial gains, outperforming AggLM and even GPT-5 in the mathematics domain.

Concurrent work PDR primarily focuses on refinement behavior to optimize the latency–accuracy trade-off, whereas PaCoRe is motivated by scaling test-time compute far beyond context limitations for general-purpose reasoning.
Despite architectural similarities, PaCoRe adopts a specialized data curation that explicitly targets instances requiring complex synthesis, rather than iterative correction or refinement. This design encourages reasoning behaviors beyond naive aggregation. 
As a result, PaCoRe exhibits a broader range of emergent behaviors, including systematic cross-checking and the synthesis of correct solutions even when all individual input messages are incorrect.

Recently, frontier proprietary systems
\cite{grok_4_heavy,  gemini_2p5_deep_think, gpt5} have demonstrated exceptional reasoning performance, reportedly via utilizing massive parallel TTC. 
While these systems showcase the power of this paradigm, their undisclosed technical details hinder broader scientific progress. 
PaCoRe bridges this gap by providing an effective, and open-sourced framework for parallel coordinated reasoning, accelerating future research in this direction.

\begin{table}[bhtp]
    \centering
    \begin{tabular}{lcc}
        \toprule
        Setting & \textbf{HMMT 2025} & \textbf{LiveCodeBench} \\
        \midrule
        $|M|=4$ & 83.6 & 63.3 \\
        $|M|=8$ & 83.7 & 64.4 \\
        $|M|=16$ & 84.2 & 64.1 \\
        $|M|\sim U(1,16)$ & 84.3 & 64.3 \\
        \textbf{$|M|\sim U(8,16)$} & \textbf{85.2} & \textbf{65.1} \\
        \bottomrule
    \end{tabular}
    \caption{
    Ablation of how the message-set size $|M|$ is determined during training.
    }
    \label{tab:sample_M_sizes}
\end{table}

\section{Conclusion and Future Work}
In this work, we introduce Parallel Coordinated Reasoning (PaCoRe), a test-time scaling framework that enables massive test-time compute (TTC) without exceeding the context window constraints of contemporary language models. 
PaCoRe addresses the limitations of sequential reasoning by shifting the primary driver of inference to coordinated parallel breadth. 
It operates via a message-passing architecture that iteratively launches many parallel reasoning trajectories, compacts their findings into context-bounded messages, and synthesizes these messages to guide subsequent exploration. 
Trained end-to-end with large-scale, outcome-based reinforcement learning, the model develops the synthesis capabilities required for effective coordination. Our empirical results demonstrate significant improvements across diverse domains. Notably, our PaCoRe-8B model achieves 94.5\% on HMMT 2025, surpassing GPT-5 by scaling effective TTC to roughly two million tokens, all orchestrated within a standard context window.

The development of PaCoRe opens several promising avenues for future research:
\begin{itemize}
    \item \textbf{Scaling to Extremes:}  We plan to further scaling PaCoRe on model sizes, task domains and further on test time compute: applying this framework to more capable foundation models,  extending the application to agentic tasks and multi-modal understanding, and expanding both the breadth (number of parallel trajectories) and depth (number of coordination rounds) to solve some extremely hard problems.
    \item \textbf{Boosting Token Intelligence Density:} While we currently scale by volume, we aim to maximize the utility of every unit of compute spent. This involves enabling more efficient parallel exploration through better organization, cooperation, and division of labor among trajectories.
    \item \textbf{Emergent Multi-Agent Intelligence:} We are interested in exploring the joint training of both the synthesis policy and the message-passing mechanism, laying minimal yet rich cooperative multi-agent learning environment, offering a valuable playground for studying emergent communication, self-organization, and collective intelligence.
    \item \textbf{Ouroboros for Pre- and Post-Training: } We intend to investigate the development of advanced synthetic data generation techniques with PaCoRe pipeline to improve both current pretraining and post-training processes.  
\end{itemize}

\section{Acknowledgements}
We express our gratitude to Song Yuan, Wuxun Xie, Mingliang Li, and Bojun Wang for their support regarding inference, and to Xing Chen, Yuanwei Lu, Changyi Wan and Yu Zhou for their assistance with training.
We also thank Shaoliang Pang, Changxin Miao, Xu Zhao, Wei Zhang, Zidong Yang, Junzhe Lin, Yuxiang Yang, Chen Xu, Xin Li and Bin Wang for their help with operational issues and platform support.
We extend our thanks to Xiaoxiao Ren, Zhiguo Huang, and Kang An for their assistance with data management support.
Special thanks go to Liang Zhao, Jianjian Sun, Zejia Weng, and Jingjing Xie for their helpful discussions throughout this project.
We acknowledge the dedicated support of the Infrastructure Team and the Data Technical Team, as well as the valuable feedback from our other colleagues at StepFun and Tsinghua University.
This work was supported by computing resources and infrastructure provided by \href{https://www.stepfun.com/}{StepFun}.

\newpage
\bibliographystyle{unsrt}

\bibliography{references.bib}
\newpage

\appendix
In this Appendix, we provide more elaboration on the implementation details, experiment results, and qualitative results.
\section{Initial Checkpoint Derivation}
To facilitate effective test-time scaling, we first implement a reasoning-oriented post-training pipeline for Qwen3-8B-Base, comprising Supervised Fine-Tuning (SFT) and Reinforcement Learning with Verifiable Rewards (RLVR).

\paragraph{Reasoning-Oriented SFT.}
We collect millions of prompts from the open-source community, spanning diverse domains including mathematics, coding, science, software engineering, tool use, logical reasoning, and creative writing. 
Leveraging these prompts, we distill data from multiple frontier models. 
Following a rigorous correctness verification process, we curate a dataset comprising 10.4M samples, totaling 61.1B tokens. 

To ensure data quality and integrity, we apply two pipes to further filter the dataset. 
First, we use predefined rules to eliminate low-quality data with degenerate patterns, such as infinite repetition, harmful content, and personally identifiable information. 
Second, we conduct comprehensive benchmark decontamination to prevent leakage. This involves both exact matching (with digit masking to catch questions featuring only numerical modifications) and $N$-gram matching ($N=64$). %
This two-pipe filtration process yields a refined dataset of 10.2M samples, totaling 59.5B tokens.

We perform SFT on Qwen3-8B-Base using the refined dataset. The training is configured with a maximum sequence length of 64k and a global batch size of 64. We employ a cosine learning rate scheduler with a 200-step warmup phase, where the learning rate peaks at $1 \times 10^{-4}$ and anneals to a final value of $1 \times 10^{-5}$. We implement domain-specific sampling weights for the dataloader, which translate to varying epochs for different domains. In total, the model is trained over $\sim$165B tokens.

\paragraph{Reasoning-Oriented RLVR.}
Similarly, we source $\sim$500k prompts paired with ground truth from the open-source community, which encompass domains including mathematics, coding, science, logical reasoning, and instruction following. We implement a robust internal RL framework utilizing vLLM for inference and Megatron for training. During each step, we sample 256 prompts and generate 16 responses per prompt (max length 64k). To address long-tail inference latency, we enable the partial rollout strategy~\cite{team2025kimi}. 

For non-coding domains, we employ LLM-as-judge (gpt-oss-120b~\footnote{https://huggingface.co/openai/gpt-oss-120b}) to verify response consistency against the ground truth. For coding tasks, we utilize sandboxes to validate code execution against test cases with soft reward. We omit KL divergence constraints (both as reward penalties and loss terms) and entropy loss during training. We use an off-policy PPO algorithm with GAE ($\gamma=1, \lambda=1$), while omitting importance sampling. Samples are split into 4 mini-batches per iteration. We set the actor and critic learning rates to $2\times 10^{-6}$ and $5\times 10^{-6}$, respectively. Furthermore, we apply the Truncated Importance Sampling ratio (threshold $C=8$) proposed by Yao et al.~\cite{yao2025offpolicy} to mitigate training-inference inconsistency issues. The entire RL phase consists of 200 training iterations.

\begin{table}[h]
    \centering
    \resizebox{0.99\linewidth}{!}{
        \begin{tabular}{l}
            \toprule
            You are given a problem and a list of reference responses. Your job is to analyze these references and provide your own response. \\
            Original Problem: \\
            \textcolor{red}{ \{\{ original\_prompt \}\} } \\
            \\
            Reference Responses: \\
            \textcolor{red}{
            \{\% for response in ref\_responses \%\}} \\
            Reference \textcolor{red}{\{\{ loop.index \}\}}: \\
            \textcolor{red}{\{\{ response \}\} }\\
            \textcolor{red}{ \{\% endfor \%\} } \\
            \\
            Now, based on the original problem and reference responses above, please provide your own comprehensive solution. \\
            \bottomrule
        \end{tabular}
    }
    \caption{\textbf{Input serialization template for PaCoRe synthesis.} We use this template to embed the current problem $x$ (denoted as \texttt{original\_prompt}) and the compact message set $M$ (denoted as \texttt{ref\_responses}) into the model's context. In the degenerate case where the message set is empty ($M = \emptyset$), this template is bypassed, and the original problem input is passed to the model unmodified.} 
    \label{tab:pacore-template}
\end{table}

\section{PaCoRe Synthesis Prompt Template}
In the \textit{Synthesis and Parallel Exploration} stage of the PaCoRe inference pipeline, we map the problem instance $x$ and the set of compact messages $M$ into a structured natural language input using a prompting function $P(x, M)$. 
The exact serialization template is provided in Table~\ref{tab:pacore-template}. By framing the compact messages as "Reference Responses," we explicitly encourage the model to critically evaluate and synthesize the diverse perspectives accumulated from the previous round.

While the method is described in Section~\ref{sec:training_details} in the context of a single-turn problem, this framework is naturally extensible to multi-turn, interactive environments (e.g., dialogues or agentic tool-use loops). 
In such settings, the pipeline is invoked at any decision point where the model must generate an action. 
The "Original Problem" slot in the template is populated by the most recent observation or user message, while the preceding interaction history (previous turns, tool outputs) remains intact in the context. 
This allows PaCoRe to maintain exactly same interface as standard chat or reasoning models, ensuring seamless compatibility with existing ecosystem.

\section{PaCoRe Training Data Preparation}
\subsection{Math Data}

\paragraph{Math Competition and Open Source Data.}
We aggregate math problems from a mix of open-source datasets and competition archives. On the dataset side, we include open source datasets from~\cite{yu2025dapoopensourcellmreinforcement,hu2025openreasonerzero,guha2025openthoughts,li2024numinamath,albalak2025bigmath,mitra2024orcamath,aslawliet2024olympiads,aslawliet2024cnk12,openr12025math220k,limo2025lessismore,muennighoff2025s1simpletesttimescaling,deepmath2025deepmath103k}. On the competition side, we collect all available historical problems (including official solutions where available) from AIME, HMMT, SMT, CMIMC, BRUMO, BMT, CHMMC, DMM, MNYMO, PUMAC, and Math Prize for Girls, using their official online archives~\cite{aime_aops_archive,hmmt_official,smt_official,cmimc_official,brumo_official,bmt_official,chmmc_official,dmm_official,mnymo_official,pumac_official,mathprize_official}. We exclude AIME 2024/2025, HMMT Feb 2025, BRUMO 2025, SMT 2025, CMIMC 2025, and HMMT Nov 2025 from the training pool because these contests are used as part of the MathArena benchmark~\cite{balunovic_srimatharena_2025}.

\paragraph{Synthetic Data.}
Beyond mined data, we add a synthetic component to strengthen large-integer arithmetic, a basic skill that underlies many competition-style solutions and naturally produces diverse numeric answers. 
We generate 13k synthetic large-integer arithmetic problems using a small set of simple hand-written text templates. For each problem, we independently sample integers \(A\) and \(B\) uniformly between \(10^{11}\) and \(10^{13}\), randomly choose one of four operation types---addition, subtraction, multiplication, or modular exponentiation modulo a fixed large prime---and substitute \(A\), \(B\), and the chosen operation into a template to form the natural-language question. We compute the corresponding exact integer answer with standard arithmetic and record it as the ground-truth label. By restricting to integer-valued outputs, each synthetic problem has a single unambiguous label and can be checked easily.

\paragraph{Data Quality Control.}
We apply several stages of quality control to ensure that the math data are well-posed and rule-checkable. First, we run deterministic rule-based filters that remove problems with embedded images or external links; multi-part questions or prompts that request multiple final answers; and items whose solutions depend on vague prose or open-ended discussion rather than a single numerical or algebraic target. 
Next, in-house math experts spot-check samples and mark common failure types (incorrect official solutions, internally inconsistent statements, or ambiguous wording). 
Based on these annotations, we construct a 100-sample evaluation set with a 1:1 mix of valid and invalid QA pairs and iteratively refine a prompt for an LLM-based judge. 
To accommodate the strict formatting requirements of our verifiable reward environment (e.g., single-part questions with a unique final answer, as in math-verify-style checkers), we use gpt-oss-120b with this prompt in a 4-pass unanimity scheme, keeping only items that receive 4/4 positive votes on the QA-validity classification. 
We tune the prompt until the F1 score on the evaluation set saturates, and in practice this configuration reliably filters out most ill-posed or non-checkable problems. 
The filtered open-source datasets and competition problems are then combined with the synthetic arithmetic problems described above. 
We summarize the problem counts from each source after rule-based and model-based filtering in Table~\ref{tab:math_data}. 
As discussed in Section \ref{sec:training_details}, we additionally apply an accuracy-based filter using a strong proposer model to select problems that are neither trivial nor degenerate; we treat this as part of curriculum design rather than basic data cleaning.

\begin{table}[htbp]
  \centering
  \begin{tabular}{lccc}
    \toprule
    Source & Before Filtering & Stage 1 & Stage 2 \\
    \midrule
    Open-source datasets & 695k & 0.7k & 0.5k \\
    Competition archives & 8.4k & 0.2k & 0.3k \\
    Synthetic arithmetic & 13.4k & 0.8k & 1.6k \\
    \midrule
    Total & 717k & 1.7k & 2.4k \\
    \bottomrule
  \end{tabular}
  \caption{Problem counts by source after rule-based and model-based filtering.}
  \label{tab:math_data}
\end{table}

\subsection{Competitive Code Data}
\paragraph{Competitive Programming Contest and Open Source Data.} About 29k problems are gathered from competitive programming problem sources (for example, subsets of TACO~\cite{li2023tacotopicsalgorithmiccode}, USACO~\footnote{https://usaco.org/index.php?page=training}, etc) and undergo a rigorous cleaning process. 
The main metrics in validation are format checks in problem statements, number of test cases, and full judging of provided code submissions. 
We use testlib~\footnote{https://github.com/MikeMirzayanov/testlib} library to help judge a submission code. 
We apply an auto-matching method on test case outputs of problems to decide a pre-defined testlib checker to judge. 
We make some modification to the checker to handling broader cases, including change `icmp` function to support 64-bit integer comparision. 
For problems requiring a special-judge checker, we use LLMs to generate it, or simply skip this problem. 
There are about 14k additional problems sampled from recent open-source competitive programming datasets, including am-thinking-v1 \cite{ji2025amthinkingv1advancingfrontierreasoning} and deepcoder \cite{deepcoder2025}. These problems are filtered with deduplication and validation processes.

\paragraph{Synthetic Data.}
To address the shortage of test cases in open-source datasets, we employ a generator-validator pipeline inspired by CodeContests+~\cite{wang2025codecontestshighqualitytestcase}. This process utilizes LLMs to produce new test cases, which are subsequently verified against ground-truth solutions and incorrect submissions. In RL experiments, we found that a fine-grained reward function based on test case pass rates significantly outperforms a simple binary reward. Our final code data source comprises approximately 5k problems curated from CodeForces~\footnote{https://codeforces.com}.

\section{More Ablation Studies}

\paragraph{Multi-round Inference Recipe.}
We study the impact of the inference trajectory configuration $\vec{K}$ in a multi-round setting. 
Fixing the first round width $K_1=32$, we ablate the second round width $K_2$ (using a configuration $\vec{K} = [32, K_2]$). Table~\ref{tab:ablation_k2} indicates that $K_2=4$ yields the best performance on both benchmarks.

\begin{table}[h]
    \centering
    \begin{tabular}{lcc}
        \toprule
        $K_2$ & \textbf{HMMT 2025} & \textbf{LiveCodeBench(2408-2505)} \\
        \midrule
        2 & 94.0 & 77.4 \\
        \textbf{4} & \textbf{94.6} & \textbf{78.4} \\
        8 & 94.0 & 77.5 \\
        \bottomrule
    \end{tabular}
    \caption{Ablation over the intermediate round width $K_2$ during inference, using a fixed multi-round configuration of $\vec{K}=[32, K_2]$.} 
    \label{tab:ablation_k2}
\end{table}

\end{document}